\documentclass[
]{ceurart}

\usepackage{listings}
\usepackage{graphicx}
\usepackage{subfigure}
\usepackage{bbding}
\usepackage{xspace}

\makeatletter
\DeclareRobustCommand\onedot{\futurelet\@let@token\@onedot}
\def\@onedot{\ifx\@let@token.\else.\null\fi\xspace}

\def\etal{\emph{et al}\onedot}
\makeatother

\begin{document}

\copyrightyear{2025}
\copyrightclause{Copyright for this paper by its authors. Use permitted under Creative Commons License Attribution 4.0 International (CC BY 4.0).}

\conference{MiGA@IJCAI25: International IJCAI Workshop on 3rd Human Behavior Analysis for Emotion Understanding, August 29, 2025, Guangzhou, China.}

\title{Online Micro-gesture Recognition Using Data Augmentation and Spatial-Temporal Attention}



\author[1]{Pengyu Liu}[%
orcid=0000-0002-3396-3108,
email=lpynow@gmail.com,
]

\author[2]{Kun Li}[%
orcid=0000-0001-5083-2145,
email=kunli.hfut@gmail.com,
]
\cormark[1]

\author[1,4,6]{Fei Wang}[%
orcid=0009-0004-1142-6434,
email=jiafei127@gmail.com,
]

\author[1,5]{Yanyan Wei}[%
orcid=0000-0001-8818-6740,
email=weiyy@hfut.edu.cn,
]

\author[3]{Junhui She}[%
orcid=0009-0004-3666-931X,
email=shejunhui@mail.ustc.edu.cn,
]

\author[1,4]{Dan Guo}[%
orcid=0000-0003-2594-254X,
email=guodan@hfut.edu.cn,
]

\address[1]{School of Computer Science and Information Engineering, School of Artificial Intelligence, Hefei University of Technology (HFUT)}
\address[2]{ReLER, CCAI, Zhejiang University, China}
\address[3]{Institute of Advanced Technology, University of Science and Technology of China}
\address[4]{Institute of Artificial Intelligence, Hefei Comprehensive National Science Center, China}
\address[5]{Key Laboratory of Knowledge Engineering with Big Data (HFUT), Ministry of Education.}
\address[6]{Xinsight Lab, Research Institute, Hefei Zhongjuyuan Intelligent Technology Co., Ltd., China}

\cortext[1]{Corresponding author.}

\begin{abstract}
In this paper, we introduce the latest solution developed by our team, HFUT-VUT, for the Micro-gesture Online Recognition track of the IJCAI 2025 MiGA Challenge. The Micro-gesture Online Recognition task is a highly challenging problem that aims to locate the temporal positions and recognize the categories of multiple micro-gesture instances in untrimmed videos. Compared to traditional temporal action detection, this task places greater emphasis on distinguishing between micro-gesture categories and precisely identifying the start and end times of each instance. Moreover, micro-gestures are typically spontaneous human actions, with greater differences than those found in other human actions. To address these challenges, we propose hand-crafted data augmentation and spatial-temporal attention to enhance the model’s ability to classify and localize micro-gestures more accurately. Our solution achieved an F1 score of 38.03, outperforming the previous state-of-the-art by 37.9\%. As a result, our method ranked \textbf{first} in the Micro-gesture Online Recognition track.
\end{abstract}


\maketitle

\section{Introduction}
When humans express emotions or interact with the world, various non-verbal forms of communication play a crucial role in the transmission of emotion and information~\cite{liu2021imigue,chen2023smg,chen2019analyze,liu2025survey,wei2020learning,gao2024dentity,guo2024benchmarking,guo2024mac,zhao2025temporal}. 
During such interactions, the human body often displays numerous spontaneous actions and gestures. Understanding these subtle behaviors~\cite{li2025prototypical,li2023data,wang2024frequency,wang2024eulermormer} is essential for gaining deeper insight into human behavior patterns and emotional states~\cite{chen2023smg,chen2019analyze}. Examples include gestures such as ``folding arms'', ``playing or adjusting hair'', and ``crossing legs''. In many scenarios, individuals may consciously suppress or conceal their genuine emotions due to social etiquette or contextual considerations. However, because micro-gestures are often spontaneous, they can serve as an indicator of a person's true emotional state.
Although there have been great successes in conventional action understanding~\cite{li2025repetitive,wang2025exploiting}, the research on micro-gesture analysis~\cite{li2023joint,chen2024prototype,liu2024micro} is still in its infancy.

Due to the significant imbalance in the category distribution of the SMG~\cite{chen2023smg} dataset, we introduced data augmentation to expand the number of samples of the rare category. At the same time, to address the issue of the model's insufficient focus on key temporal and spatial information for micro-gesture recognition, we designed a spatial-temporal attention module to strengthen the ability to recognize key areas. In summary, the main contributions of this paper are as follows:
\begin{itemize}
\item We introduce a spatial-temporal attention to enhance the baseline’s localization head, encouraging the model to focus on more informative areas of the feature output. Compared with the baseline model, our approach achieves improved action classification and more precise boundary localization.

\item To mitigate the severe category imbalance caused by the spontaneous nature of micro-gestures in real-world scenarios, we augment the dataset to improve the model’s sensitivity to gesture categories with fewer samples. This improves the classification performance for gesture categories with fewer samples.

\item In the Micro-gesture Online Recognition challenge, our proposed method achieved an F1 score of \textbf{38.03} in the test set, securing first place in the competition. Experimental results demonstrate that our model is capable of effectively distinguishing and localizing micro-gestures\protect\footnotemark[1].

\footnotetext[1]{The Kaggle competition page: \href{https://www.kaggle.com/competitions/the-3rd-mi-ga-ijcai-challenge-track-2/leaderboard}{https://www.kaggle.com/competitions/the-3rd-mi-ga-ijcai-challenge-track-2/leaderboard}}
\end{itemize}

\section{Related Work}

\subsection{Micro-Gesture Analysis Datasets}
The SMG~\cite{chen2023smg} dataset is designed for studying naturally occurring micro-gestures under stress. It contains micro-gestures collected from 40 participants of diverse ages, genders, and ethnic backgrounds. The dataset categorizes micro-gestures into 16 categories and has been widely used in micro-gesture recognition and emotion analysis tasks, demonstrating its practicality and effectiveness in these research areas. The iMiGUE~\cite{liu2021imigue} dataset is the first publicly available dataset aimed at recognizing and understanding suppressed or hidden emotions through micro-gestures. It includes 359 videos with a total duration of 2,092 minutes, collected from 72 subjects from 28 countries. 
Some studies suggest that micro-actions~\cite{guo2024benchmarking,li2025prototypical,gu2025motion,sun2023unified}, which focus on spontaneous actions of the whole body, can better reflect subtle emotional changes in humans. The MA-52~\cite{guo2024benchmarking} dataset consists of 52 micro-action categories and 7 body part labels, covering a wide range of natural micro-actions. It comprises 22,422 instances collected from 205 participants during psychological interview sessions. 
In addition, Li~\etal~\cite{li2024mmad} introduced the Multi-label Micro-Action 52 (MMA-52) dataset and proposed a Multi-label Micro-Action Detection task, which aims to recognize all micro-actions within a video sequence for fine-grained understanding.
The MMA-52 dataset comprises 6,528 videos and 19,782 action instances collected from 203 subjects.

\subsection{Micro-gesture Online Recognition}
Guo~\etal~\cite{guo2023micro} proposed a novel deep network that integrates Graph Convolutional Networks (GCNs) and Transformers to extract motion features from 2D skeleton sequences. This hybrid design leverages the strengths of both GCNs and Transformers, effectively capturing spatial relationships and long-range temporal dependencies. Their method achieved first place in the Micro-gesture Online Recognition track of the MiGA 2023 Challenge. Wang~\etal~\cite{wang2024micro} developed a deep network with dual-stream input for micro-gesture online recognition. They first used a sequential action recognition model to extract gesture features from RGB and skeleton sequences, respectively, and then used a multi-scale Transformer encoder to process these features as a detection model. Their approach secured first place in the Micro-gesture Online Recognition track of the MiGA 2024 Challenge. Additionally, Liu~\etal~\cite{liu2024micro} proposed a model using learnable query points and Mamba blocks. They used learnable points to learn the positions of frames that are more important for gesture recognition and leveraged Mamba's ability to efficiently capture complex relationships in sequence data, significantly improving the model's ability to recognize micro-gestures, and achieved second place in the MiGA 2024 challenge using only RGB data.

\subsection{Temporal Action Detection}
Temporal Action Detection (TAD)~\cite{tan2022pointtad,liu2024end,zhang2022actionformer,yang2024dyfadet} aims to locate and classify all actions in untrimmed videos. Existing methods can generally be divided into two categories: feature-based approaches and end-to-end approaches. Feature-based methods typically rely on pre-trained feature extractors to obtain video representations, which are then used for subsequent processing. In contrast, end-to-end approaches jointly optimize video encoders and decoders to achieve better task-specific feature representations. End-to-end approaches allow more seamless modeling through the simultaneous optimization of both encoding and decoding stages. For example, Tan~\etal~\cite{tan2022pointtad} proposed an end-to-end action detection model, PointTAD, which utilizes learnable query points to accurately localize and differentiate actions in videos. Liu~\etal~\cite{liu2024end} introduced the concept of fine-tuning large language models into the TAD task by employing VideoMAE~\cite{tong2022videomae, wang2023videomae} as the backbone and fine-tuning it for action localization, achieving precise classification and localization. On the other hand, feature-based approaches are favored for their efficiency and lower computational cost. Tirupattur~\etal~\cite{tirupattur2021modeling} incorporated an attention-based multi-label dependency layer into their model, significantly improving the modeling of co-occurrence and temporal dependencies between actions. Dai~\etal~\cite{dai2022ms} proposed a novel ConvtransFormer network that effectively integrates global and local temporal relations. Zhang~\etal~\cite{zhang2022actionformer} employed a Transformer encoder to capture long-range dependencies, while Shi~\etal~\cite{shi2023tridet} introduced a ternary point modeling approach for more accurate boundary localization. Yang~\etal~\cite{yang2024dyfadet} dynamically aggregated multi-scale features to handle actions of varying temporal lengths. Together, these approaches provide valuable insights into accurately localizing and distinguishing complex temporal actions in untrimmed videos.

\section{Methodology}

\subsection{Problem Definition}

Given an untrimmed video $V$, represented as a sequence of feature vectors $V = \{v_1, v_2, \ldots, v_T\}$, where $T$ denotes the temporal length, each $v_t \in V$ is typically extracted using pre-trained video encodes, such as I3D~\cite{carreira2017quo} or VideoMAE~\cite{tong2022videomae, wang2023videomae}. The objective of Micro-gesture Online Recognition is to predict a set of action instances $\Psi = \{\psi_1, \psi_2, \ldots, \psi_n\}$, where $N = \{1, 2, \ldots, n\}$ denotes the number of predicted instances. Each instance $\psi_n = \{t_s^n, t_e^n, c_n\}$ is defined by its start time $t_s^n$, end time $t_e^n$, and category label $c_n$, where $c_n \in C$, and $C$ is the set of all predefined action categories.

\subsection{Data Augmentation}

\begin{figure}[t!]
\centering
\includegraphics[width=0.9\linewidth]{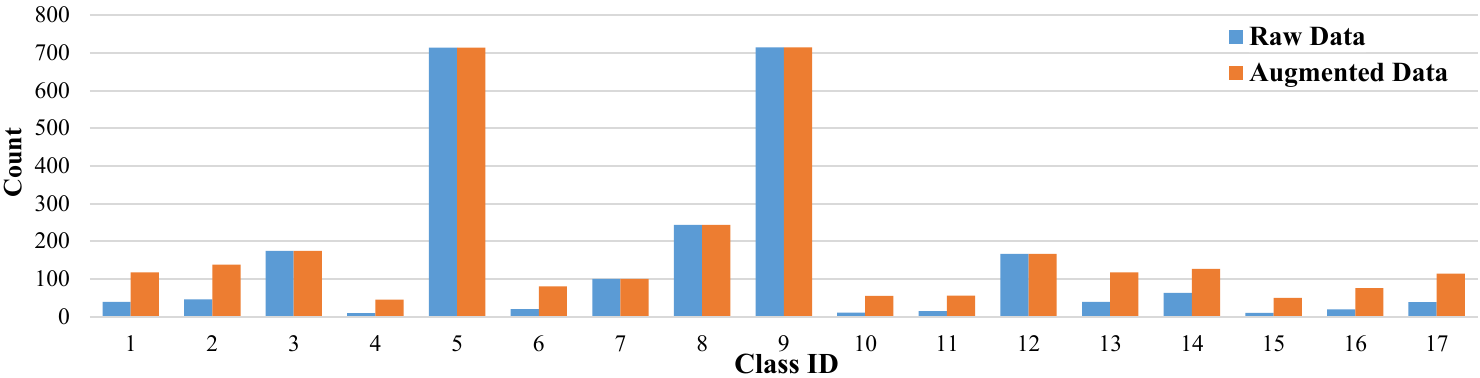}
\caption{Micro-Gesture Category Distribution on the training set of the SMG dataset.  
The blue side is the SMG dataset, while the orange side is the data after applying data augmentation. The number of samples in underrepresented categories has significantly increased following augmentation.
}
\label{fig:data_aug}
\end{figure}

The SMG dataset captures micro-gestures that naturally occur in daily life. However, the frequency of occurrence varies between different gesture categories. For instance, gestures like ``Moving legs'' appear far more frequently than rarer gestures such as ``Touching or covering suprasternal notch.'' 
To address the severe category imbalance in the training set, we designed a category-frequency-based adaptive label augmentation strategy. 

First, we count the number of instances for each gesture category in the training set and define a minimum instance threshold $\alpha$. For categories with fewer than $\alpha$ instances, we calculate the number of times each instance needs to be repeated using a logarithmic function:
\begin{equation}
R_c = \left\lfloor \log_2\left(\frac{\alpha}{Z_c}\right) \right\rfloor + 1,
\label{eq:data_aug}
\end{equation}
where $\alpha$ is the minimum instance threshold, and $Z_c$ is the number of instances of category $c$ in the training set. 
Specifically, for each instance, if its $Z_c < \alpha$, we consider it a rare category. We then replicate its annotations in the training data according to the calculated $R_c$, effectively increasing the representation of that category. This augmentation is performed at the annotation level rather than duplicating raw video data, preserving both the structure and diversity of the dataset. It avoids redundancy caused by naive duplication and enhances the model's ability to learn rare categories.
Figure~\ref{fig:data_aug} demonstrates the distribution of instances across the gesture category before and after applying our data augmentation strategy. It clearly shows a significant reduction in category imbalance, thus enabling the model to better learn rare gestures during training.

\subsection{Overall Architecture}

\begin{figure*}[t!]
\centering
\includegraphics[width=1.0\linewidth]{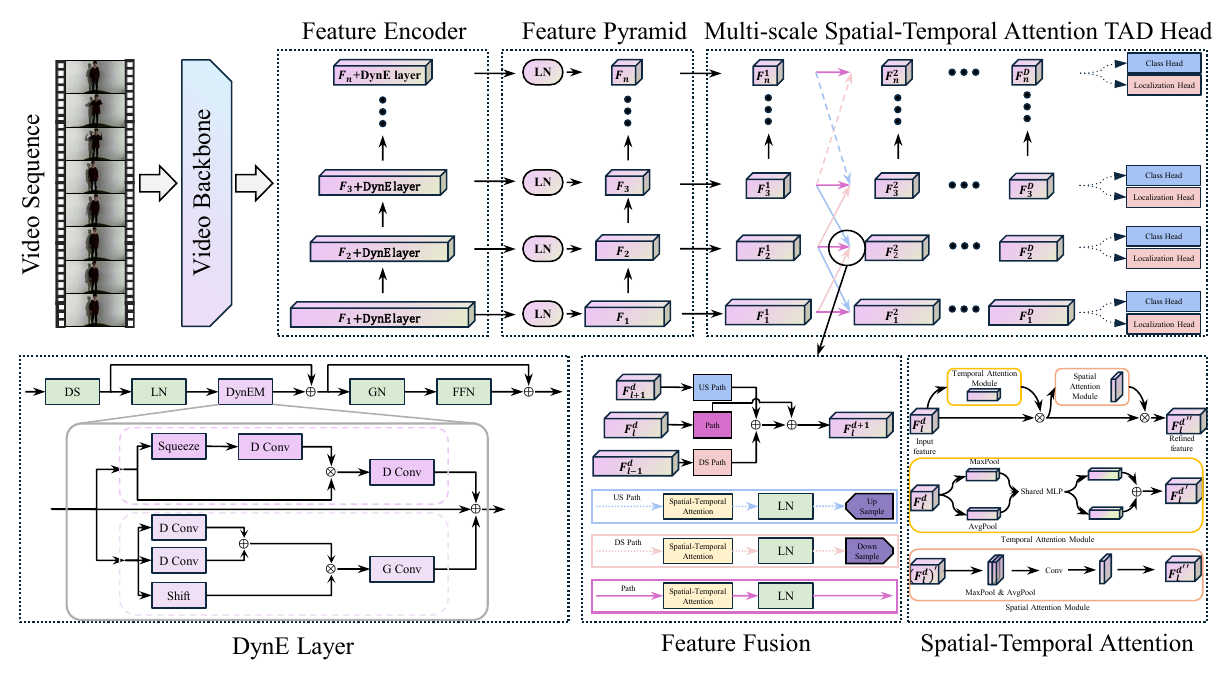}
\caption{Overview of the proposed method. The Feature Encoder consists of downsampling and DynE layers. The multi-scale fusion module in the Multi-scale Spatial-Temporal Attention TAD Head comprises Feature Fusion and Spatial-Temporal Attention.}
\label{fig:overview}
\end{figure*}

Considering the unique characteristics of micro-gestures, we enhance the DyFADet~\cite{yang2024dyfadet} to build a micro-gesture online recognition model that extracts discriminative representations from video encoder features and dynamically adjusts the detection head for actions of varying lengths. Following the DyFADet model architecture, our approach is composed of three main components: a feature extraction backbone, an encoder, and a multi-scale spatial-temporal attention TAD head. Specifically, we first use a pretrained video encoder to extract video features. These features are then passed through an encoder to generate a feature pyramid. Within this pyramid, the features are downsampled using a stride of 2 via the Dynamic Encoder (DynE) layer to obtain representations at different temporal scale features $\textbf{\textit{F}}_n$, where $n = 1, 2, \ldots, N$, and $N$ is the total number of pyramid levels. Finally, the multi-scale spatial-temporal attention TAD head is used to detect action categories and their temporal boundaries. The overall architecture is illustrated in Figure~\ref{fig:overview}. 

The feature encoder in DyFADet introduces the DynE to enhance both global and local modeling capabilities during action detection. DynE is designed based on the Transformer, replacing standard self-attention with DynE. It contains two parallel branches: the instance-dynamic branch and the multi-kernel branch. These branches collaboratively generate dynamic weighted masks to improve the discriminative power of the feature representations. In the instance-dynamic branch, a DFA convolution with kernel size 1 is applied to model features and generate a global temporal attention mask that captures overall action information. In contrast, the multi-kernel branch applies DFA convolutions with multiple kernel sizes to generate attention masks with various receptive fields, better adapting to local structural diversity. The two branches are defined as:
\begin{equation}
\textbf{\textit{F}}_{\text{instance-dynamic}} = \text{DFA\_Conv}_1 \left(\text{Squeeze}(\text{LN}(\text{DS}(\textbf{\textit{F}})))\right),
\label{eq:instance}
\end{equation}
\begin{equation}
\textbf{\textit{F}}_{\text{multi-kernel}} = \text{DFA\_Conv}_{k, w} \left(\text{LN}(\text{DS}(\textbf{\textit{F}}))\right),
\label{eq:kernels}
\end{equation}
where DS denotes the down-sample the feature with a scale of 2 to generate the representations with different temporal resolutions. LN is Layer Normalization, Squeeze represents average pooling along the channel dimension, and $w$ is a parameter used to expand the convolution window size for better temporal modeling. The outputs of both branches are then added to the original input features, forming the final representation of the DynE. In the complete feature encoding process, each DynE performs downsampling on the input features, constructing a multi-scale temporal feature representation. This dynamic feature selection mechanism effectively mitigates the lack of feature discriminability in previous models and significantly improves the overall action detection performance.

\subsection{Spatial-Temporal Attention}

To address the inconsistency in detecting short-duration and long-duration instances caused by traditional static methods, shared detection heads, DyFADet~\cite{yang2024dyfadet} adopts a Multi-Scale TAD Head architecture. This architecture dynamically fuses features across multiple scales to guide the detection head in adaptively adjusting its parameters based on the input context. However, micro-gestures exhibit strong spatial-temporal dependencies, and the current Multi-Scale TAD Head design lacks sufficient attention to spatial-temporal features, which may lead to inaccurate boundary localization. Therefore, we introduce a Spatial-Temporal Attention to enhance the model's sensitivity to spatial-temporal information. As illustrated in the figure~\ref{fig:overview}, our detection head takes input features from the current level of the pyramid along with its neighboring upper and lower levels. Through the Spatial-Temporal Attention module, along with upsampling (US) and downsampling (DS) operations, we construct three parallel paths for subsequent detection tasks.

For the Spatial-Temporal Attention module, in the Temporal Attention branch, we first compress the input feature map along the spatial dimensions to obtain a one-dimensional vector. During spatial compression, both average pooling and max pooling are considered. These operations aggregate the spatial information of the feature maps and are fed into a shared MLP network to generate a temporal attention map. The spatially compressed features are summed element-wise to yield the final temporal attention, which is defined as:
\begin{equation}
Attn_T(\textbf{\textit{F}}) = \sigma(\text{MLP}(\text{AvgPool}(\textbf{\textit{F}})) + \text{MLP}(\text{MaxPool}(\textbf{\textit{F}}))),
\label{eq:temporal}
\end{equation}
where $\sigma$ denotes the sigmoid function.

In the Spatial Attention branch, the input features are compressed along the channel dimension using both average pooling and max pooling. The resulting pooled features are concatenated and passed through a convolutional layer to produce the spatial attention map. This can be formulated as:
\begin{equation}
Attn_S(\textbf{\textit{F}}) = \sigma(Conv^{7 \times 7}([\text{AvgPool}(\textbf{\textit{F}}); \text{MaxPool}(\textbf{\textit{F}})])),
\label{eq:spatial}
\end{equation}
where $\sigma$ is the sigmoid function and $7 \times 7$ denotes the convolution kernel size.

Finally, our classification and boundary regression modules operate on the aggregated features from all pyramid levels. The class head uses a 1D convolution followed by a sigmoid function to predict the probability of each action category at each time. The localization head applies a ReLU-activated 1D convolution to estimate the temporal offsets from the current time to the start and end times of the action instance. This unified structure enables robust detection of actions with varying durations while incorporating both temporal and spatial information, thereby achieving more generalized and accurate temporal action localization.

\section{Experiments}

\subsection{Dataset and Evaluation Metric}
The SMG~\cite{chen2023smg} dataset consists of 3,692 samples covering 17 micro-gesture categories. A cross-subject evaluation protocol is adopted, wherein 40 subjects are divided into two groups: a training group comprising long video sequences from 35 subjects, and a testing group consisting of sequences from the remaining 5 subjects. 
The dataset provides both RGB data and skeleton data. However, we achieve state-of-the-art performance using only RGB data as input. We jointly evaluate the detection and classification performance of algorithms using the F1 score:
\begin{equation}
F1 = \frac{2 \cdot \text{Precision} \cdot \text{Recall}}{\text{Precision} + \text{Recall}}.
\label{eq:F1}
\end{equation}

Given a long video sequence for evaluation, Precision is the ratio of correctly classified micro-gestures to the total number of gestures retrieved by the algorithm in the sequence. Recall is the ratio of correctly retrieved micro-gestures to the total number of annotated micro-gestures in the ground truth. This metric comprehensively reflects the algorithm’s ability to both detect and correctly classify micro-gestures.

\subsection{Implementation Details}
We use VideoMAEv2-g~\cite{tong2022videomae} as the video backbone to extract features from video sequences. The videos are processed at the original frame rate of 28 fps, and a sliding window mechanism is adopted for feature extraction, where each window contains 16 frames with a stride of 4 frames. To standardize the input size of the model, all frames are resized to 224×224. 
The batch size is set to 128, the initial learning rate is 1e-4, and the training is conducted for a total of 400 epochs.

\subsection{Experimental Results}

\textbf{Main Comparison.} 
The experimental results comparing the performance of our method with other models are shown in Table~\ref{tab:Table1}. On the test set of the SMG dataset, our method using VideoMAEv2-g features surpassed the previous state-of-the-art performance, and our team's model ranked first. Our solution achieved an F1 score of \textbf{38.03}, outperforming the previous state-of-the-art by 37.9\%. The proposed data augmentation and spatial-temporal attention demonstrate high performance, proving their effectiveness in micro-gesture online recognition and indicating their ability to capture richer semantic features.

\begin{table}[htbp]
\centering
\caption{The top-4 results of Micro-gesture Online Recognition on the SMG test set. Data is provided by the Kaggle competition page\protect\footnotemark[1].
}
\begin{tabular}{c|cc}
\toprule
Rank  & Team  & F1 Score \\
\midrule
1     & \textbf{HFUT-VUT (Ours)} & \textbf{38.03} \\
\midrule
2     & Chutian Meng & 31.53 \\
\midrule 
\midrule
 & NPU-MUCIS (IJCAI 2024 MiGA2 Rank 1)~\cite{wang2024micro} & 27.57 \\
\midrule
 & HFUT-VUT (IJCAI 2024 MiGA2 Rank 2)~\cite{liu2024micro} & 14.34 \\
\midrule
 & Guo~\etal (IJCAI 2023 MiGA2 Rank 1)~\cite{guo2023micro} & 14.85 \\
\bottomrule
\end{tabular}
\label{tab:Table1}
\end{table}
\footnotetext[1]{The Kaggle competition page: \href{https://www.kaggle.com/competitions/the-3rd-mi-ga-ijcai-challenge-track-2/leaderboard}{https://www.kaggle.com/competitions/the-3rd-mi-ga-ijcai-challenge-track-2/leaderboard}}

\textbf{Ablation Studies.} In Table~\ref{tab:baseline}, we report the results of our experiments conducted on several baselines, comparing our proposed method with existing approaches to demonstrate its effectiveness. We adopted VideoMAEv2-g as the backbone and applied the method proposed by Liu~\etal~\cite{liu2024end}, achieving the best result with an F1 score of \textbf{38.03}. However, due to time constraints, we did not further optimize or fine-tune the model, and we report this result solely as a reference for future research. Furthermore, we illustrate the effectiveness of the proposed data augmented strategy and the Spatial-Temporal Attention. We conducted the following experiments: a baseline experiment, an experiment incorporating data augmentation, one incorporating Spatial-Temporal Attention, and one combining both techniques. 
The performance of all experiments outperformed the baseline. Since data augmentation effectively addressed the category imbalance of micro-gestures, enabling the model to focus more on and distinguish rare categories, performance improved by 15.33\%. For the detection head, we find that using spatial-temporal attention in the detection head can better leverage information, improving performance by 17.69\%. However, using data augmentation or spatial-temporal attention alone only provides limited improvements. When combining data augmentation and spatial-temporal attention, our method achieves a performance of \textbf{33.44}, significantly outperforming the baseline.

\begin{table}[]
\centering
\caption{F1 Scores of Baseline and Our Model on the SMG Dataset.}
\begin{tabular}{c|c|c|c|c}
\hline
\textbf{Data Augmentation} & \textbf{Spatial-Temporal Attention} & \textbf{Method} & \textbf{Backbone} & \textbf{F1 score} \\ \hline
\XSolidBrush & \XSolidBrush & AdaTAD~\cite{liu2024end}       & VideoMAE-Small & 17.95          \\
\XSolidBrush & \XSolidBrush & AdaTAD~\cite{liu2024end}       & VideoMAE-Base  & 18.72          \\
\XSolidBrush & \XSolidBrush & AdaTAD~\cite{liu2024end}        & VideoMAEv2-g  & 38.03          \\
\XSolidBrush & \XSolidBrush & ActionFormer~\cite{zhang2022actionformer}  & VideoMAEv2-g     & 15.25          \\
\XSolidBrush & \XSolidBrush & DyFADet~\cite{yang2024dyfadet}       & VideoMAEv2-g     & 27.78          \\ \hline
\CheckmarkBold & \XSolidBrush & Ours & VideoMAEv2-g & 32.04 \\
\XSolidBrush & \CheckmarkBold & Ours & VideoMAEv2-g & 32.69 \\
\CheckmarkBold & \CheckmarkBold & Ours & VideoMAEv2-g & \textbf{33.44} \\ \hline
\end{tabular}
\label{tab:baseline}
\end{table}

\subsection{Error Analysis}
In addition, to evaluate the performance of our proposed model, we followed the standard practice in action detection by using the action detection evaluation toolkit proposed by Alwassel~\etal~\cite{alwassel2018diagnosing}.

\begin{figure}[htbp]
\centering
\subfigure[False Negative Analysis on Baseline]{
\includegraphics[width=0.42\textwidth]{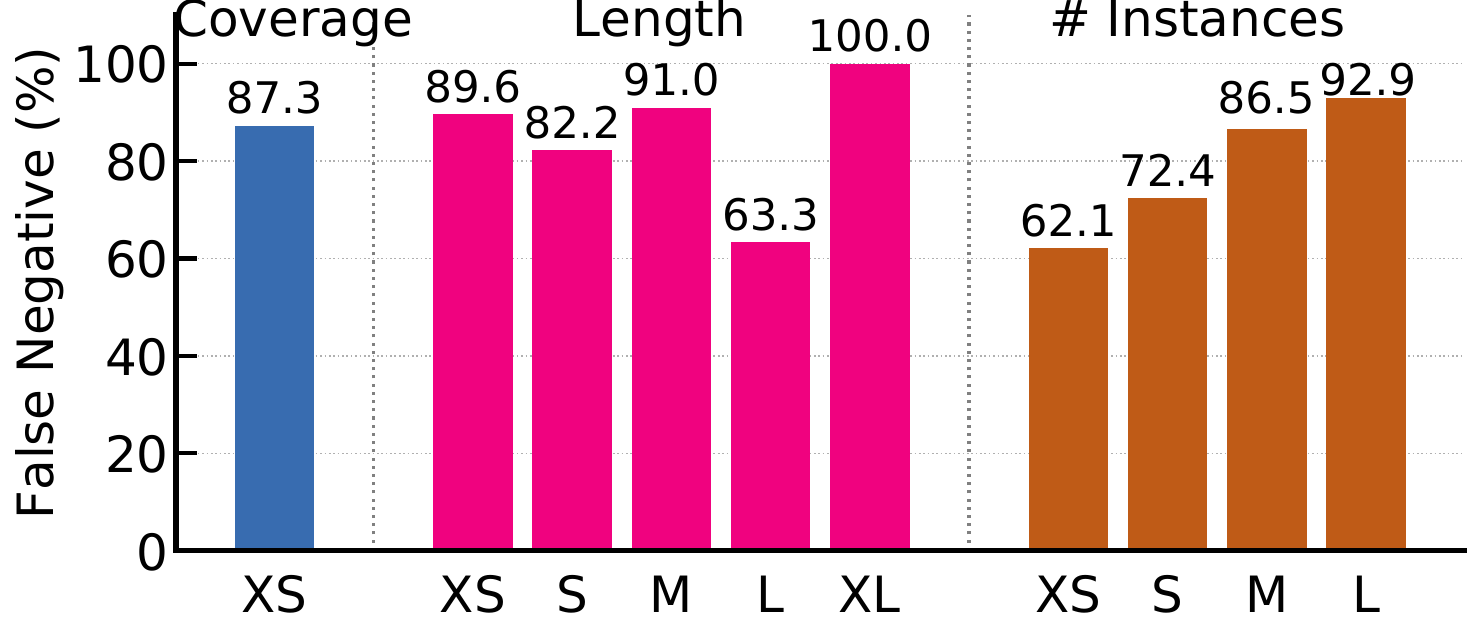}
\label{fig:detad-a}
}
\hspace{0.4em}
\subfigure[False Negative Analysis on Our Method]{
\includegraphics[width=0.42\textwidth]{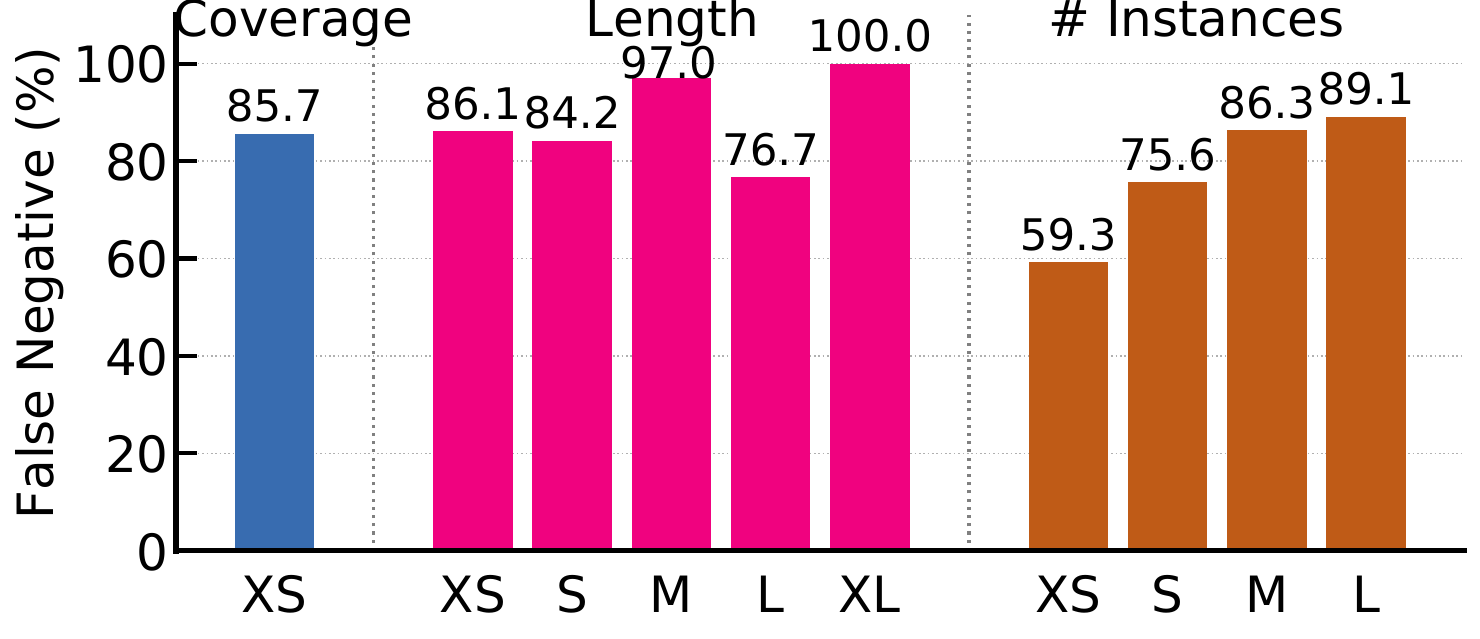}
\label{fig:detad-b}
}
\subfigure[False Positive Analysis on Baseline]{
\includegraphics[width=0.46\textwidth]{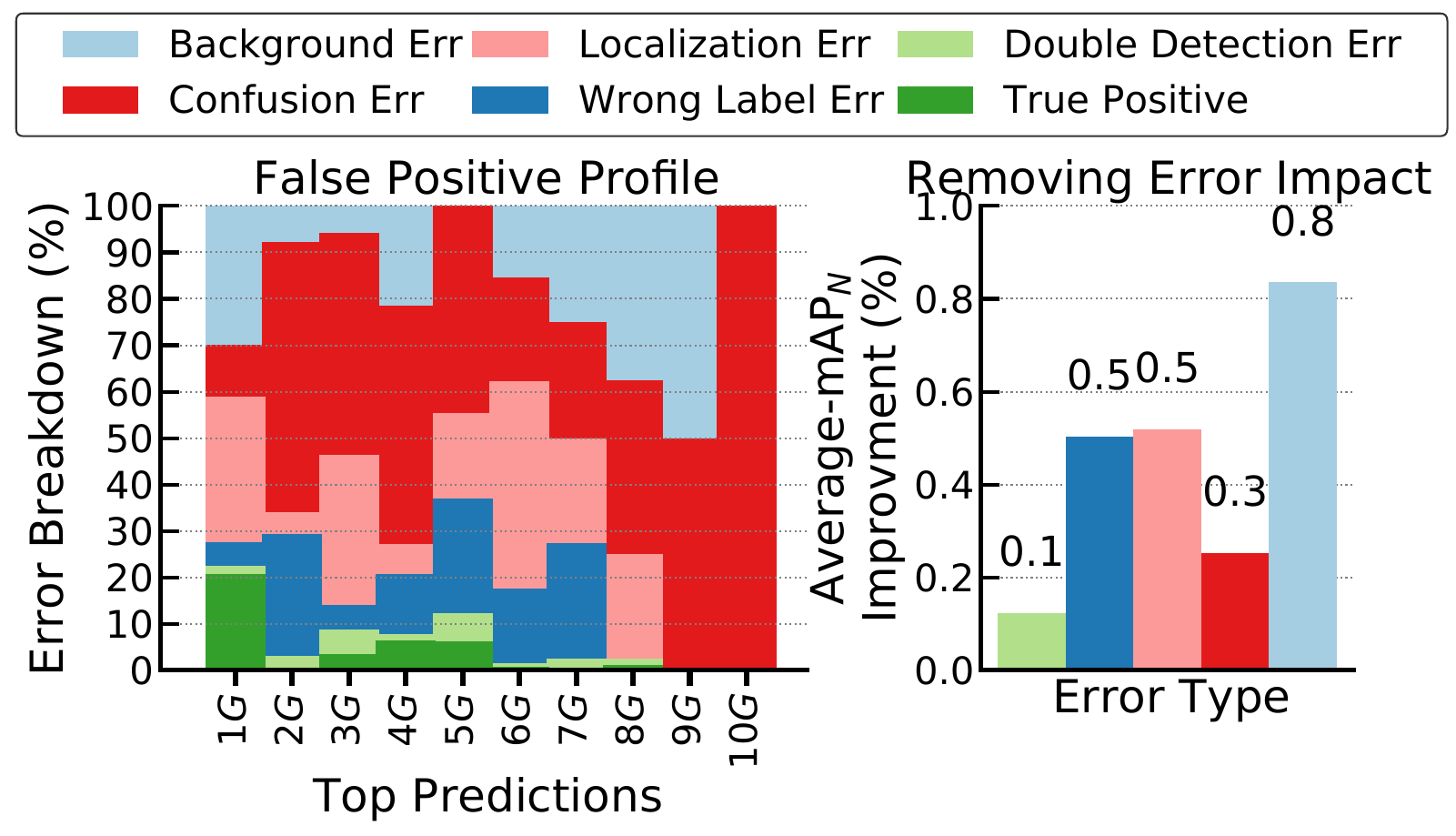}
\label{fig:detad-c}
}
\subfigure[False Positive Analysis on Our Method]{
\includegraphics[width=0.46\textwidth]{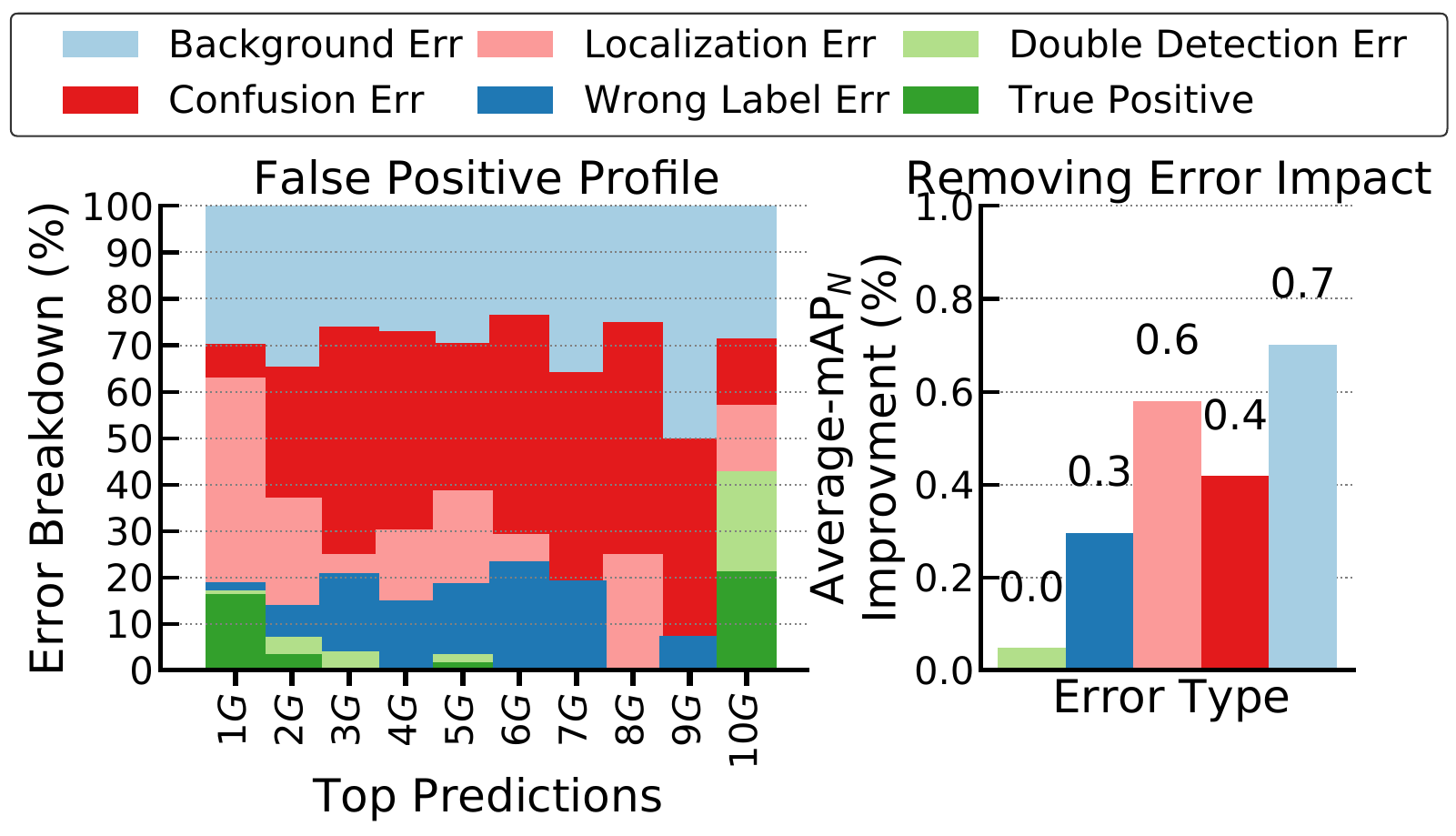}
\label{fig:detad-d}
}
\subfigure[Sensitivity Analysis on Baseline]{
\includegraphics[width=0.68\textwidth]{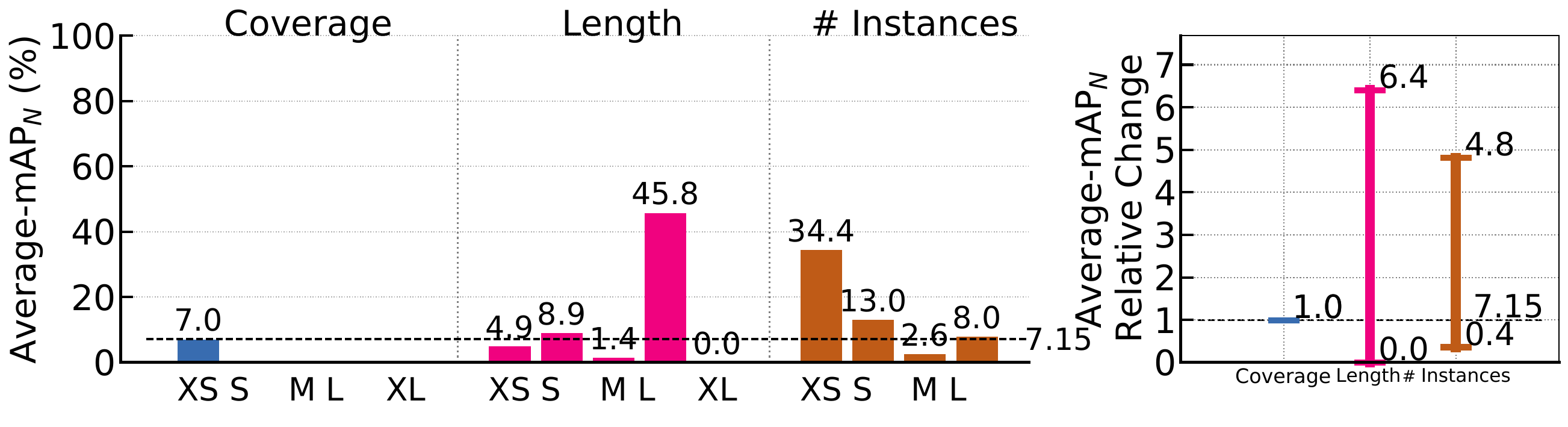}
\label{fig:detad-e}
}
\subfigure[Sensitivity Analysis on Our Method]{
\includegraphics[width=0.68\textwidth]{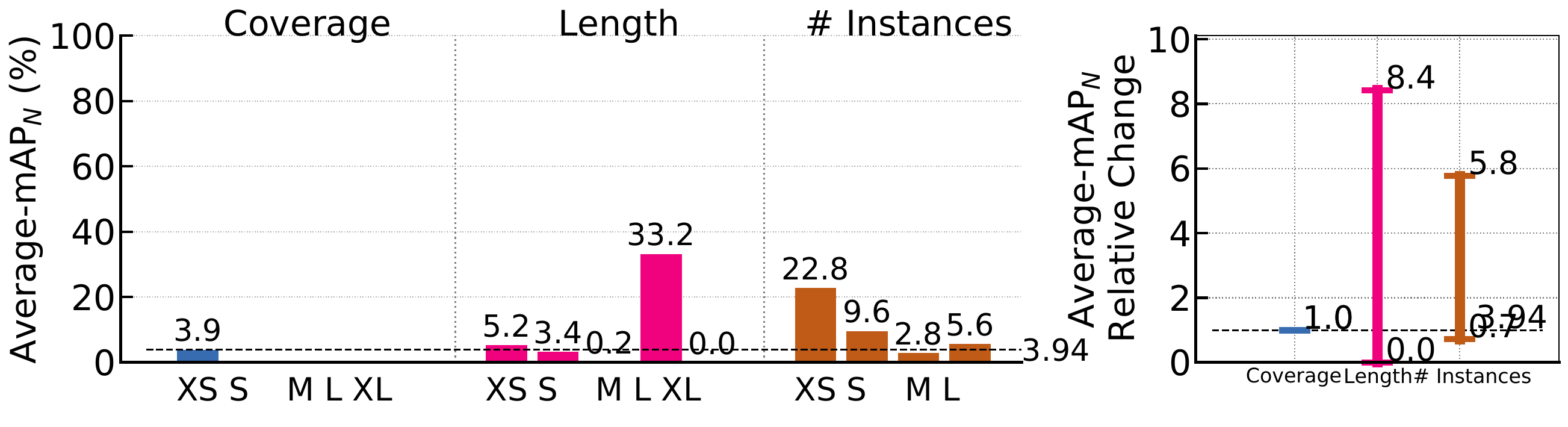}
\label{fig:detad-f}
}
\caption{Error Analysis of the baseline and the proposed method using a standard TAD evaluation toolkit~\cite{alwassel2018diagnosing}. Figures (a), (c), and (e) show the performance metrics of the baseline, while Figures (b), (d), and (f) illustrate the corresponding results achieved by our method.
}
\label{fig:6figs}
\end{figure}

\begin{figure*}[t!]
\centering
\includegraphics[width=1.0\linewidth]{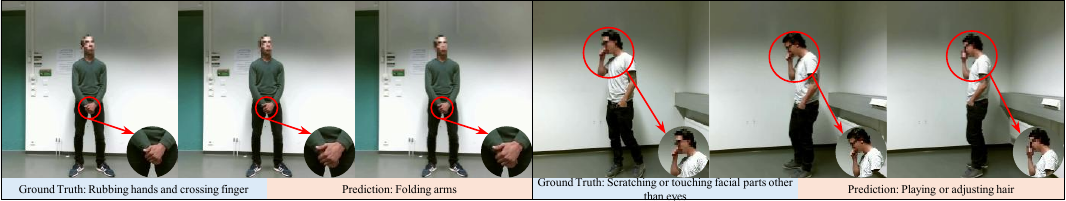}
\caption{Typical examples of classification errors. Left figure: The model incorrectly predicted ``Rubbing hands and crossing fingers'' as ``Folding arms''; Right figure: The model incorrectly predicted ``Scratching or touching facial parts other than eyes'' as ``Playing or adjusting hair''.}
\label{fig:falsedetection}
\end{figure*}

\textbf{False Negative Analysis.} Figures~\ref{fig:detad-a} and~\ref{fig:detad-b} illustrate the false negative analysis for the baseline and our improved methods, focusing on the true action instances that were not detected by the models. By analyzing the missed detection rates at a tIoU threshold of 0.5 under different Coverage, Length, and Instance conditions, we can evaluate the performance of both models in various scenarios. To align with the characteristics of the SMG dataset, we define the Length intervals as [0, 2, 5, 7, 9.75, INF] and the Instance intervals as [-1, 15, 100, 200, INF], which are labeled on the axes as [XS, S, M, L, XL]. It can be observed that our model significantly reduces the false negative rate for short-duration actions, which account for the majority of the data. Additionally, in low-density videos, our model shows a lower rate of missed detections compared to the baseline. In summary, the false negative analysis indicates that our model demonstrates stronger detection capability on samples with a higher number of action instances.

\textbf{False Positive Analysis.} Figures~\ref{fig:detad-c} and~\ref{fig:detad-d} show the false positive analysis of the baseline and our improved method, focusing on five common types of false detection errors. We present the false positive analysis at tIoU = 0.5, where the x-axis represents the top NG predictions, with G denoting the number of ground truth instances. From the comparison, it can be observed that false positives are mainly concentrated in confusion errors and background errors. Compared to the baseline, our method significantly reduces confusion errors. Moreover, for nearly all categories of removing error impact, our model achieves a clear reduction, indicating improvements not only in boundary localization accuracy but also in action classification precision. In summary, the false positive analysis indicates that our proposed approach achieves improvements across nearly all top predictions. At the 10G level, our model shows the most significant reduction in confusion errors, demonstrating a substantial enhancement in the model’s ability to distinguish between different actions.

\textbf{Sensitivity Analysis.} 
Finally, we analyzed the model’s sensitivity to variations in action characteristics. As shown in Figures~\ref{fig:detad-e} and~\ref{fig:detad-f}, compared to the baseline, our method exhibits a noticeably reduced sensitivity to changes in coverage, length, and instance count. In particular, the sensitivity drop is more significant for actions with length category L and instance count category XS. This indicates that our model achieves stronger robustness and generalization in micro-gesture online detection.

Figure~\ref{fig:falsedetection} shows typical examples of incorrect micro-gesture classification by the model. The left figure shows the model incorrectly predicting ``Rubbing hands and crossing fingers'' as ``Folding arms,'' and the right figure shows the model incorrectly predicting ``Scratching or touching facial parts other than eyes'' as ``Playing or adjusting hair.'' These two incorrect predictions demonstrate the model's lack of sensitivity to finger movements and its difficulty in classifying micro-gestures involving multiple body parts, such as limbs and the head.

\section{Conclusion}

In this paper, we presented our solution for the Micro-gesture Online Recognition track of the IJCAI 2025 MiGA Challenge. Our approach is based on the DyFADet model, with the introduction of a data augmentation strategy to alleviate the severe category imbalance commonly observed in real-world micro-gestures. Additionally, we enhance the model’s ability to capture the spatial-temporal dependencies of micro-gestures by incorporating a Spatial-Temporal Attention into the detection head. Our final model achieved a score of 38.03 on the test set of the SMG dataset. Notably, our model relies solely on RGB data for recognition. In future research, we plan to explore a wider variety of data augmentation methods to address the issue of overfitting that may be caused by data augmentation based on category frequency, thereby improving the model's generalization. At the same time, we plan to incorporate skeleton data and explore joint modeling using both RGB and skeleton modalities to further improve online micro-gesture recognition performance.

\begin{acknowledgments}
This work is supported by National Key R\&D Program of China (NO.2024YFB3311602), Natural Science Foundation of China (62272144), the Anhui Provincial Natural Science Foundation (2408085J040), and the Major Project of Anhui Provincial Science and Technology Breakthrough Program (202423k09020001), and the Fundamental Research Funds for the Central Universities (JZ2024HGTG0309, JZ2024AHST0337).
\end{acknowledgments}

\bibliography{sample-ceur}

\end{document}